\documentclass[fleqn,10pt]{wlscirep}
\usepackage[utf8]{inputenc}
\usepackage[T1]{fontenc}
\usepackage{float}
\usepackage{placeins} 
\usepackage{multirow}
\usepackage{geometry}

\usepackage{etoolbox} 

\AtBeginEnvironment{algorithm}{%
  \setlength{\linewidth}{\dimexpr\textwidth-2cm\relax}
  \SetAlgoNlRelativeSize{-1}
  \SetNlSty{textbf}{}{}
  \SetAlgoNlRelativeSize{0}
  \DontPrintSemicolon
  \setlength{\algomargin}{1em}
  \SetAlCapHSkip{0pt}
  
}

\usepackage{hyperref}
\hypersetup{
    colorlinks=true,
    linkcolor=black, 
    citecolor=black, 
    urlcolor=black   
}
\usepackage{array}
\usepackage{tabularx}
\usepackage{inputenc}
\usepackage[linesnumbered,ruled,vlined]{algorithm2e}
\usepackage[linesnumbered,ruled,vlined]{algorithm2e}
\SetAlgoNlRelativeSize{-1} 
\SetNlSty{textbf}{}{} 
\SetAlgoNlRelativeSize{0} 
\DontPrintSemicolon 
\SetAlCapHSkip{0pt} 
\usepackage{amsthm,cite,url,graphicx,booktabs,lipsum,color,bm,caption,subcaption,soul}

\DeclareUnicodeCharacter{FF0C}{\,}

\title{Developing an AI-based Integrated System for Bee Health Evaluation}

\author{Andrew Liang}

\affil{alcago2020@gmail.com}


\begin{abstract}

Honey bees pollinate about one-third of the world's food supply, but bee colonies have alarmingly declined by nearly 40\% over the past decade due to several factors, including pesticides and pests. Traditional methods for monitoring beehives, such as human inspection, are subjective, disruptive, and time-consuming. To overcome these limitations, artificial intelligence has been used to assess beehive health. However, previous studies have lacked an end-to-end solution and primarily relied on data from a single source, either bee images or sounds. This study introduces a comprehensive system consisting of bee object detection and health evaluation. Additionally, it utilized a combination of visual and audio signals to analyze bee behaviors. An Attention-based Multimodal Neural Network (AMNN) was developed to adaptively focus on key features from each type of signal for accurate bee health assessment. The AMNN achieved an overall accuracy of 92.61\%, surpassing eight existing single-signal Convolutional Neural Networks and Recurrent Neural Networks. It outperformed the best image-based model by 32.51\% and the top sound-based model by 13.98\% while maintaining efficient processing times. Furthermore, it improved prediction robustness, attaining an F1-score higher than 90\% across all four evaluated health conditions. The study also shows that audio signals are more reliable than images for assessing bee health. By seamlessly integrating AMNN with image and sound data in a comprehensive bee health monitoring system, this approach provides a more efficient and non-invasive solution for the early detection of bee diseases and the preservation of bee colonies.

\end{abstract}
\begin{document}
\flushbottom
\maketitle
\section*{Introduction}

Honeybees are one of the most important pollinators, and they contribute approximately \$500 billion annually to the global food industry. In the United States, honey bees provide pollination services valued at \$15 to \$20 billion annually in 2020 \cite{Bush:2020}. However, bee populations in the United States have declined by nearly 40\% over the past decade\cite{Steinhauer:2023}, posing severe risks to agricultural productivity and food availability.\\

\noindent Beekeeping faces significant challenges, with colony decline as a primary concern. A significant contributor to this decline is colony collapse disorder (CCD), where bees abandon their hives. Although the exact causes of the disorder remain unclear, potential reasons include pesticides, pests, and malnutrition. Pesticides used in agriculture pose direct and indirect threats to bees, affecting both the insects and their food sources. Varroa mites spread viruses, and insufficient nutrition reduces bee vitality \cite{Haydak:1970}. The challenges extend to urbanization and climate changes, impacting bee foraging and nesting activities. Addressing these interconnected challenges is essential for the preservation of pollination services and the health of ecosystems. 
\\

\noindent Beekeepers perform a crucial response to the challenges outlined by monitoring bee health to ensure that the factors that cause CCD are absent. By tracking bee populations, a diverse range of hive behaviors, and overall colony health, beekeepers can identify potential threats and take appropriate action. Traditional approaches, such as human inspection and bee sampling, offer valuable insights into colony health but can be subjective, intrusive, and labor-intensive. As a result, these approaches may not capture subtle changes in bee behavior, causing delayed detection of health
issues. In recent years, alternative methods such as sensor technology and remote monitoring have provided a non-disruptive way to obtain continuous information on bee activities and environmental conditions. However, these advancements may not fully realize their potential in assessing apiary well-being, as they primarily offer basic monitoring functions without delivering deeper insights into hive health.\\ 

\noindent With the advancements in machine learning, researchers are pushing the boundaries of innovation in beekeeping. Computer vision is applied to analyze bee images, enabling the detection of changes in bee movement, posture, and interactions that might escape the human eye \cite{spiesman2021assessing, bhuiyan2022artificial, sledevivc2018application, ratnayake2021tracking, knauer2022bee, liang2022effectiveness, kaur2022cnn}. Additionally, the analysis of audio recordings from beehives extracts and scrutinizes various audio features, such as the frequency and intensity of bee sounds, providing valuable signals about the colony's activity level, queen presence, and even predicting swarming events\cite{zgank2018acoustic, kim2021acoustic, ruvinga2021lstm, orlowska2021queen}. The integration of these sophisticated computer vision and signal processing techniques into Internet of Things (IoT) systems helps effectively pinpoint signs of stress, disease, or other health issues \cite{hong2020long,
zacepins2016remote, gil2017honey, ntawuzumunsi2021self}. However, current research lacks a unified solution that fully integrates bee object detection and health assessment. Furthermore, there is a lack of in-depth exploration into the potential synergies from combining diverse data sources. Thus, a unified system that consolidates these diverse data sources is crucial to help beekeepers .\\

The study addresses these challenges with two main innovations. First, it introduces a framework capable of identifying bees and assessing their health. This framework employs advanced techniques for the precise detection of bees in images and sounds, ensuring that only data containing bees is considered for subsequent health assessment. Secondly, the study combines visual and audio signals to more effectively analyze bee behavior. A special machine learning model, the Attention-based Multimodal Neural Network (AMNN),  was designed to adaptively focus on the most relevant features from images and sounds, significantly enhancing bee health assessments. The comprehensive approach could be implemented in advanced remote beekeeping tools, offering beekeepers deeper insights for more effective hive management and supporting bee conservation.\\
    
\section*{Methods}

The proposed beehive health monitoring system comprises the following crucial steps, as illustrated in Figure \ref{fig:workflow}:

\begin{figure}[tb]
    \centering  \includegraphics[width=1\textwidth, height=5cm]{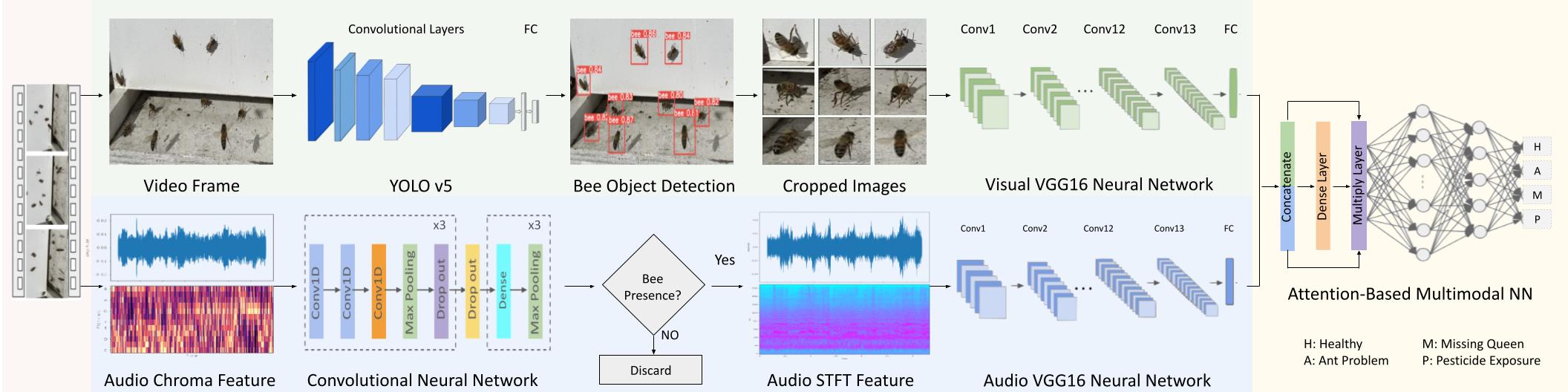}
    \caption{Beehive health 
    monitoring system workflow: The video footage of bee activity is first captured by a camera and microphone. Then, bees are identified and cropped from the images, and bee sounds are extracted from audio clips. Next, both visual and audio features are extracted using the VGG16 model. Subsequently, these signals are concatenated and processed by the attention-based multimodal neural network (AMNN) for bee health assessment.}
    \label{fig:workflow}
\end{figure}

\begin{itemize}[itemsep=-2pt]
    \item Data acquisition and annotation, building the foundation for subsequent analysis
    \item Data preprocessing, including audio feature extraction
    \item Bee object detection,  identifying bees in images and audio clips
    \item Bee health assessment, using either bee visual or audio signals
    \item Comprehensive bee health assessment, synergizing both visual and audio signals
\end{itemize}

\subsection*{Data Acquisition}
This study generated four distinct datasets for bee object detection and health assessment, incorporating both image and audio data for a comprehensive analysis. Video recordings were conducted at the entrances of 25 beehives located in three apiaries across San Jose, Cupertino, and Gilroy in California, USA. The data collection, spanning from October 2022 to March 2023, captured the activities of an estimated 20,000 to 30,000 bees in each hive. An extensive 50-hour recording of bee activities resulted in the collection of a total of 150 gigabytes of video data. \\

\noindent Four health labels were established to examine variations in bee health, including healthy bees, those from colonies with missing queens, affected by ant infestations, or exposed to pesticides. These conditions were introduced in a controlled environment to simulate specific challenges in two hives. These challenges impacted the entire colony, enabling a comprehensive assessment of their effects. Each condition was applied for five days to gather representative data. Trials for each condition were conducted with a minimum of a four-week interval, providing the colonies time for recovery.\\

\noindent To capture images, an Arducam IMX519 Raspberry Pi camera was positioned above beehive entrances. The camera had a back-illuminated stacked sensor with a pixel size of \(1.22 \, \mu\text{m} \times 1.22 \, \mu\text{m}\) and an autofocus lens with an aperture of \(f/1.75\), enabling it to capture focused and high-resolution images, as well as record videos in 720p at 60 frames per second. The camera was placed to ensure that bees entering or leaving the hive would be observed. \\

\noindent In addition to visual data, audio was captured using a high-quality PoP voice Professional microphone positioned near the beehives. With a high audio sensitivity of \(30 \, \text{dB}\) and a frequency response of \(20 \, \text{Hz}\) to \(20 \, \text{kHz}\), the microphone ensured that sounds of various frequencies produced by the bees and their surroundings were captured. Noise cancellation and a windscreen feature minimized background noise, further enhancing the data quality.

\subsection*{Data Annotation}

\subsubsection*{Annotation for Bee Image Object Detection}

Images were extracted from recorded videos at a frequency of one frame per second. Subsequently, the extracted images were randomly downsampled to create a representative and manageable dataset.\\

\noindent The image annotation process was conducted using the Label Studio platform. For each image, individual bees were manually annotated by drawing bounding boxes (BBox) around their bodies and wings using the provided annotation tool. The output label files were generated in the YOLO format, which recorded the BBox coordinates for each annotated bee. 
The label file contained the object ID, X-axis center, Y-axis center, BBox width, and height. All values were normalized to the image size, ranging from 0 to 1. In cases when multiple bees were present in an image, each bee was represented by a separate line in the label file.\\

\noindent To ensure consistency and accuracy, only images satisfying the following conditions were included in the final dataset:

\begin{itemize}[itemsep=-4pt]
    \item Bee visibility: Each bee image at least showed 50\% of the bee's body to ensure sufficient visual information for analysis.
    \item Image quality: Bee images that were excessively blurry or unclear were excluded to maintain data quality.
\end{itemize}

\noindent During the labeling process, bees were labeled as completely as possible. Additionally, two annotators randomly cross-checked the annotations, ensuring that the datasets contained detailed and reliable annotations.

\subsubsection*{Annotation for Bee Audio Object Detection}

Audio recordings were normalized to a uniform 10-second duration for consistency. During annotation, the strength and length of bee sounds were compared with other background noises, such as those from birds, airplanes, and cars. Audio clips were labeled as bee-related if bee sounds dominated the recording. Each recording was carefully listened to twice for enhanced accuracy.

\subsubsection*{Annotation for Bee Health Classification}

\noindent To enable subsequent bee health classification, images, and audio clips were annotated based on prior knowledge of bee health indicators. Each image and audio clip was assigned one of four categorical labels, representing distinct hive conditions. These categories included healthy bees, beehives with ant issues, missing queen, and pesticide exposure. The final datasets contain various data types and serve multiple purposes, as summarized in Table \ref{tab:dataset_summary}.

\begin{table}[tb]
    \renewcommand{\arraystretch}{.9} 
    \centering
    \begin{tabular}{lccc}
    
        \toprule
        \textbf{Data Type} & \textbf{Purpose} & \textbf{Label} & \textbf{Count} \\
        \midrule
        Images & Bee Object Detection & Bee/No Bee & 1,524 \\
        \midrule
        Audio Clips (10s each) & Bee Object Detection & Bee/No Bee & 2,840 \\
        \midrule
        \multirow{4}{*}{Images \& Audio Clips} & \multirow{4}{*}{Health Assessment} & Healthy & 1,060 \\
        & & Ant Infestation & 926 \\
        & & Missing Queen & 878 \\
        & & Pesticide Exposure & 648 \\
        \bottomrule
    \end{tabular}
\caption{Summary of bee object detection and health assessment datasets}
    \label{tab:dataset_summary}
\end{table}

\subsection*{Audio Feature Extraction}
After loading the audio samples and capturing them at a frequency of 44,100 Hz for 10 seconds, four audio features were extracted to analyze the acoustic characteristics for subsequent bee identification and health classification tasks. The extracted features included Mel Spectrogram, Mel-Frequency Cepstral Coefficients (MFCC), Short-Time Fourier Transform (STFT), and Chromagram.\\

\noindent \textbf{Mel Spectrogram} divides the audio signal into frequency bands and measures the energy in each band over time. It helps effectively capture frequency patterns associated with different bee behaviors. This makes it particularly suitable for analyzing bee sounds, as bees produce distinct sounds within specific frequency ranges during activities like wing beating. Mathematically, the mel-scale can be represented as:

\begin{equation}
\text{Mel}(f) = 2595 \cdot \log_{10}\left(1 + \frac{f}{700}\right)
\end{equation}

where:
\begin{align*}
& \text{Mel}(f): \text{The mel-scale value corresponding to frequency } f \\
& f : \text{The frequency in Hz}
\end{align*}

\noindent \textbf{MFCC} are coefficients that represent the short-term power spectrum of an audio signal, widely utilized in speech and audio processing. They provide valuable insights into other features relevant to various activities such as queen piping or worker piping. In this study, the MFCC was calculated using 12 Mel cepstrum coefficients through Mel-scale frequency filtering of the bee audio signal. Mathematically, the calculation of a coefficient is represented as:

\begin{equation}
c_i = \sum_{n=1}^{N_f} (S_n \cdot \cos(i(n-0.5)\cdot \frac{\pi} {N_f}))
\end{equation}

where:
\begin{align*}
& c_i: \text{The } i\text{-th MFCC coefficient} \\
& N_f: \text{The number of triangular filters in the filter bank} \\
& S_n: \text{The log energy output of the } n\text{-th filter coefficient}
\end{align*}

\noindent \textbf{STFT} transforms a time-domain signal into its frequency representation over short time intervals. This method provides a high-resolution representation of the sound, allowing for a thorough analysis of how the sounds evolve over a period of time. In this work, the STFT was computed using a window length of 1024 and a hop length of 512. The STFT value at frequency \( f \) and time \( t \) can be expressed as:

\begin{equation}
    \text{STFT}(f, t) = \int_{-\infty}^{\infty} x(t) \cdot w(t - \tau) \cdot e^{-j2\pi f \tau} \, d\tau
\end{equation}

where:
\begin{align*}
 & {STFT}(f, t) : \text{The STFT value at frequency } f \text{ and time } t \\
 & x(t) : \text{The original audio signal} \\
 & w(t - \tau) : \text{The window function centered at time } t \\
 & e^{-j2\pi f \tau} : \text{The complex sinusoidal at frequency } f
\end{align*}

\noindent \textbf{Chromagram} represents the pitch content of a sound signal by dividing the spectrum into pitch classes and measuring the energy within each pitch class over time. It is useful for the model to recognize pitch variations in bee sounds linked to specific activities such as foraging, queen piping, or signaling distress.\\

\begin{figure}[tb]
    \centering
    \includegraphics[width=.8\textwidth, height=8.3cm]{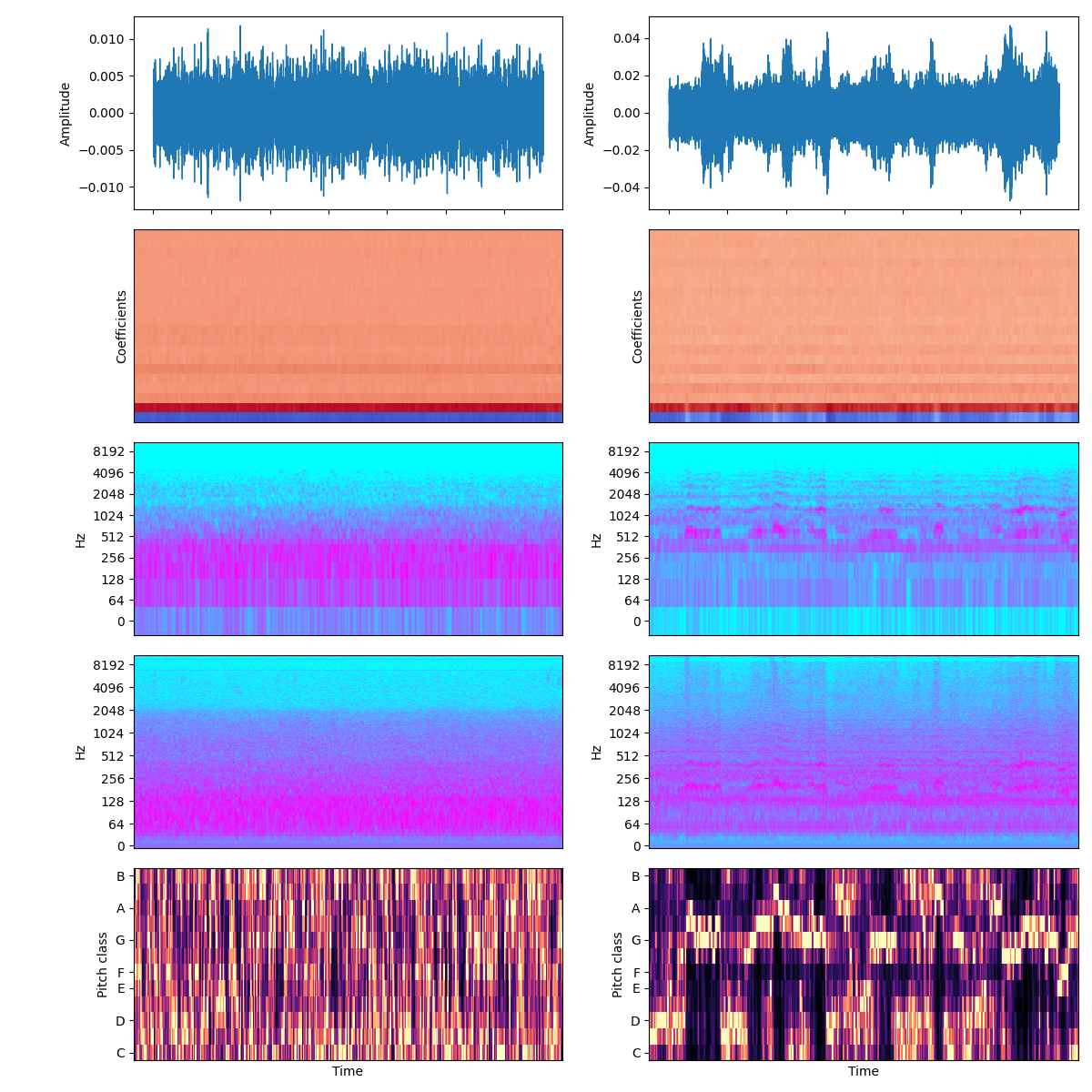}
    \caption{The audio features show different patterns when bees are absent (left) vs. present (right). The top-to-bottom representation includes the original audio wave, Mel Spectrogram, MFCC, STFT, and Chromagram.}
    
    \label{fig:audio_features}
\end{figure}

\noindent After converting the audio clips into features, these features displayed distinct patterns across various scenarios. Figure \ref{fig:audio_features} provided a visual representation of the audio features that correspond to the absence or presence of bees. In this study, the average value along the feature rows for each audio sample was calculated during audio feature extraction. Subsequently, a power-to-decibel conversion was executed, and the resulting decibel values were normalized to a range of [0, 1]. These steps effectively condensed the information and captured the essential content of the audio signals. \\

\subsection*{Model Development}
This section focuses on model training for bee object detection and health prediction, utilizing bee visual and auditory data. During model development, datasets were randomly split into three subsets: 80\% for training, 10\% for validation, and 10\% for testing. Various model algorithms and hyperparameters were assessed, and the top-performing ones were detailed below.

\subsubsection*{Bee Image Object Detection Models}
\noindent A YOLOv5 model \cite{yolov5} was developed for the localization and cropping of bees in images. This model was optimized to balance detection accuracy with computational efficiency. Various data augmentation techniques, including blurring, grayscale conversion, and contrast enhancement, were systematically applied to enhance the training dataset diversity and improve model generalization. Optimal performance was achieved by fine-tuning the hyperparameters, which included setting the model to 50 epochs, using a batch size of 64, employing Stochastic Gradient Descent (SGD) as the optimizer, and using a learning rate of 0.01 with a momentum of 0.937. The finalized YOLOv5 model consisted of 276 layers and 35,248,920 trainable parameters. The output cropped bee image would be saved for bee health evaluation.

\subsubsection*{Bee Audio Object Detection Models}
\noindent Four 1D CNN models were developed to identify bees in audio clips. Each model used a distinct audio feature: Mel Spectrogram, MFCC, STFT, or Chromagram. The algorithm began with two convolutional layers, each with 64 filters and an 8-unit kernel for capturing bee sound frequencies.  Next, Batch normalization was implemented to expedite model training. A max-pooling layer was then added to reduce spatial data dimensions, followed by a dropout layer with a rate of 25\% to mitigate overfitting. The pattern was repeated twice: first with two convolutional layers of size 128, then with another two convolutional layers of size 256, each followed by a batch normalization, max pooling, and dropout layer. \\

\begin{figure}[b]
    \centering
    \includegraphics[width=.9\textwidth, height=3.2cm]{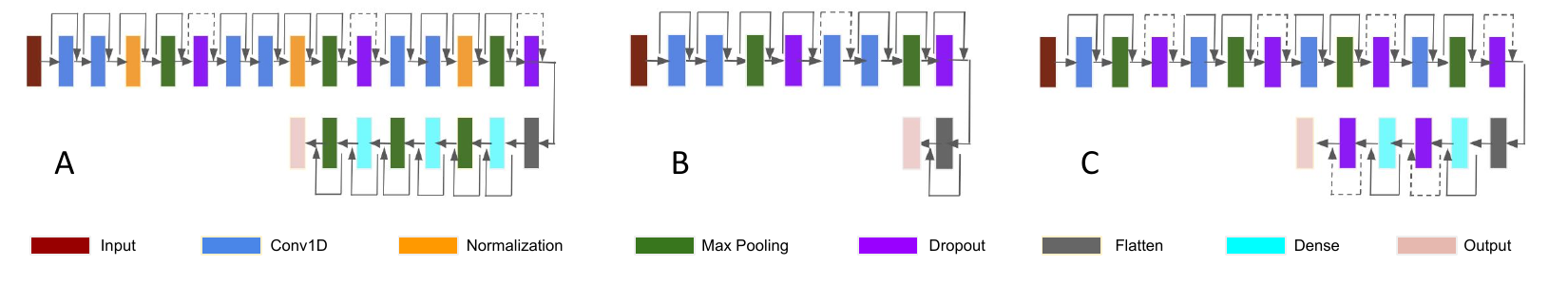}
    \caption{
Compressed self-designed CNN structures for (A) audio object detection, (B) visual health assessment, and (C) audio health assessment}
    \label{fig:cnn}
\end{figure}

\noindent Next, the data were flattened and passed into three dense layers with a size of 32, 64, and 128 neurons, respectively, to further improve feature recognition. A 25\% dropout layer followed each dense layer. The model ended with a 2-unit dense layer using softmax activation, classifying sounds as either the bee's absence or presence. The model's algorithm was shown in Figure  \ref{fig:cnn} (A). After iterative tuning, the set of hyperparameters was selected to achieve optimal model performance. The chosen values were 20 epochs, a batch size of 64, an early stopping patience of 5, a learning rate of 0.0001, and the Adam optimizer. 

\subsubsection*{Bee Image Health Assessment Models}
The visual bee health evaluation focused on classifying cropped bee images into various health categories. In this step, four distinct deep learning models were implemented as outlined below.

\begin{figure}[tb]
    \centering
    \includegraphics[width=.8\textwidth, height=4cm]{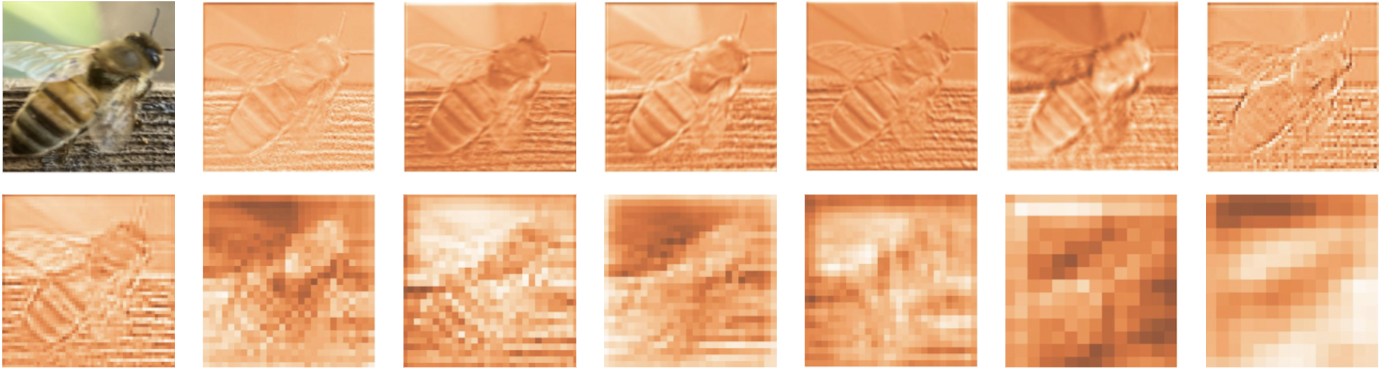}
    \caption{Visualization of VGG16's feature maps, highlighting key regions of an input image across the convolutional layers}
    \label{fig:feature_maps}
\end{figure}

\begin{itemize}[itemsep=4pt]
    \item \textbf{Inception v3 \cite{szegedy2015going}} is a pre-trained deep CNN that captures image features similar to how our brain comprehends pictures. In the study, the top layers of the base model were excluded and replaced with custom layers. The output of the base model was flattened into a 1D feature vector. Next, the vector went through a dense layer with 256 units, activated by a rectified linear unit (ReLU) for capturing non-linear patterns.  dropout layer prevented overfitting by randomly deactivating 20\% of neurons.  The model ended with a dense layer of four nodes, using softmax activation to generate class probabilities.

    \item \textbf{MobileNet v2 \cite{howard2017mobilenets}} is a pre-trained lightweight CNN known for its efficiency in mobile applications. The model uses depth-wise separable convolutions to reduce the number of parameters, making it suitable for resource-constrained environments. In this study, the top layers of the model were removed and the model's weights were frozen. Then, the same custom layers used in Inception v3 were added.

    \item \textbf{VGG16 \cite{simonyan2014very}} is a pre-trained deep CNN renowned for its robustness in various visual recognition challenges. It contains 16 layers, including 13 convolutional layers and 3 fully connected layers. In this study, the model was tailored to bee health classification by excluding the top layers and adding the same custom layers as those previously described in Inception v3 and MobileNet v2.
  
     \item \textbf{A 2D CNN} was specifically designed for the study's task. It started with two convolutional layers with 64 and 128 filters, using a 3x3 kernel and ReLU activation. These layers aimed to detect unique image features. Following this, a 2x2 max pooling layer reduced the spatial dimensions of the feature maps while retaining key details. Then, a 25\% dropout was applied to prevent overfitting.  Next, two extra convolutional layers with 256 and 1024 filters were included, along with additional max pooling and dropout layers. These layers allowed the model to capture more intricate patterns in the data. The resulting feature maps were flattened and flowed into a dense layer using softmax activation. This final layer predicted probabilities for the four bee health classes. The model structure was presented in Figure \ref{fig:cnn} (B).
     
\end{itemize}
\noindent The feature maps, as shown in Figure \ref{fig:feature_maps}, were generated within the VGG16 architecture for extracting features hierarchically from input images. These maps, generated by convolutional layers, progressively captured complex visual patterns as the network deepened. From rudimentary edges in early layers to complex shapes in deeper ones, these feature maps enhanced the network's capability to distinguish and classify visual content effectively.\\

\noindent The hyperparameters that demonstrated optimal performance in bee audio object detection also proved effective for bee visual health assessment models. As a result, these same hyperparameters were applied to train all bee health assessment models, including audio and multimodal variations. This approach ensured a fair and unbiased comparison of model performance by minimizing any potential impact of different hyperparameter configurations.

\subsubsection*{Bee Audio Health Assessment Models}

In addition to 1D CNN and VGG16 models, 2D CNN and LSTM models were also developed for bee health classification through audio analysis.

\begin{itemize}[itemsep=4pt]

    \item \textbf{The 2D CNN} architecture consists of four convolutional layers with filter sizes of 16, 32, 64, and 128, respectively. These layers extracted features from the input audio at varying levels of detail. After each convolutional layer, the features went through a 2x2 max pooling layer to reduce the feature map size, along with a 25\% dropout layer to improve generalization. The flattened output was then fed into two dense layers, each with 32 and 16 neurons. These layers used the ReLU activation for introducing nonlinearity to the network. A 25\% dropout layer followed each dense layer to enhance model robustness. The final layer, featuring four nodes with softmax activation, produced predictions for the classification task. The
model's design was illustrated in Figure \ref{fig:cnn} (C).

    \item \textbf{LSTM\cite{hochreiter1997long}} captures temporal dependencies and long-term patterns in sequential data. In the study, a custom LSTM architecture was designed with a 128-unit LSTM layer, followed by two dense layers of 64 and 32 units. Both dense layers were activated by the ReLU function to capture non-linear data patterns. Dropout layers were added to deactivate 40\% of neurons after each dense layer. The final layer included four nodes with softmax activation, representing the classification classes.
    
\end{itemize}

\subsubsection*{Attention-based Multimodal Neural Network}

Previous models for bee health assessment focused solely on either bee images or audio. To address the limitations of these isolated approaches, the AMNN was proposed to merge both bee visual and auditory information. By incorporating an attention mechanism, the model could dynamically focus on crucial features in each modality. This adaptability enabled a comprehensive understanding of bee behavior and improved bee health assessment accuracy. \\

\noindent The AMNN Framework, as illustrated in Figure 1 and Algorithm 1, can be summarized as follows:
\begin{itemize}[itemsep=-2pt]
    \item \textbf{VGG16 Feature Extractor}: VGG16 models were employed to extract hierarchical representations from input bee image and audio. The data went through a series of convolutional and pooling layers, followed by a dense layer with 16 units and a ReLU activation function. This resulted in 4,096 intermediate features for each data type.
    \item \textbf{Concatenation}: The visual and audio outputs from the dense layers were concatenated, creating a unified feature set.
    \item \textbf{Attention Mechanism}: The concatenated data passed through a dense layer with four units and a softmax activation function. The layer learned weights for each feature, indicating their respective importance. The output was flattened and then passed to a multiply layer, where attention weights were element-wise multiplied with the concatenated input. This allowed the model to focus on the most crucial features in each mode.
    \item \textbf{Fully Connected Layer}: Two additional dense layers with 32 and 16 units further refined the feature importance representation, extracting deeper insights into feature relevance. Each dense layer was followed by a dropout layer with a 50\% rate to prevent overfitting.   

    \item \textbf{Final Output Layer}: The final prediction utilized a softmax-activated output layer with four nodes. This corresponded to the four bee health conditions evaluated through image and audio analysis.
   
\end{itemize}

\begin{center}
\begin{algorithm}[H]
\SetAlgoLined
\KwResult{Predicted bee health condition}
\textbf{Function} AMNN\_Model(image\_input, audio\_input)\\
\Indp  
feature\_image $\gets$ VGG16(image\_input)\;
feature\_audio $\gets$ VGG16(audio\_input)\;
feature\_concatenated $\gets$ concatenate(feature\_image, feature\_audio)\;
attention\_layer $\gets$ Dense(4, \texttt{softmax})(feature\_concatenated)\;
feature\_attended $\gets$ multiply(attention\_layer, feature\_concatenated)\;
fc\_layer1 $\gets$ Dense(32, \texttt{relu})(feature\_attended)\;
fc\_layer1 $\gets$ Dropout(0.5)(fc\_layer1)\;
fc\_layer2 $\gets$ Dense(16, \texttt{relu})(fc\_layer1)\;
fc\_layer2 $\gets$ Dropout(0.5)(fc\_layer2)\;
output $\gets$ Dense(4, \texttt{softmax})(fc\_layer2)\;
\Return{output}\;  
\caption{Pseudocode for the AMNN model}
\end{algorithm}
\end{center}



\noindent Throughout the model training process, the weights were learned to minimize the model's loss function, which consists of two main components: the reconstruction loss for both the image and sound modalities. This loss function was formulated as a weighted combination of these two terms, as illustrated below:

\[
L = \lambda_{\text{image}} \cdot L_{\text{image}}(y, \hat{y}_{\text{image}}) + \lambda_{\text{sound}} \cdot L_{\text{sound}}(y, \hat{y}_{\text{sound}})
\]

\[
L_{\text{image}}(y, \hat{y}_{\text{image}}) = - \sum_{i=1}^{N} \sum_{j=1}^{M} y_{i,j} \cdot \log(\hat{y}_{\text{image}, i, j})
\]

\[
L_{\text{sound}}(y, \hat{y}_{\text{sound}}) = - \sum_{i=1}^{N} \sum_{j=1}^{M} y_{i,j} \cdot \log(\hat{y}_{\text{sound}, i, j})
\]

Where:
\begin{align*}
& L : \text{The loss function} \\
& y : \text{The true label for the $j$-th record in class $i$.} \\
& \hat{y} : \text{The predicted label for the $j$-th record in class $i$}\\
& \lambda: \text{The weight for the $j$-th record in class $i$} \\
& N: \text{The number of classes}\\
& M: \text{The number of records in each class} 
\end{align*}

\section*{Results}

\noindent 
The following section presented the model results in two key tasks: identifying bees from pictures and sounds, and assessing bee health. The evaluation covered both visual and audio data, as well as their combination. To ensure the accuracy of the study, a method called 5-fold cross-validation was used, where the data was randomly split into five equal parts. In one iteration, one part was used for testing and the other four for training. This process was repeated five times, with each time using a different part for testing. The study's success was measured using standard evaluation metrics, such as accuracy, precision, recall, and the F1-score, as explained below:

\begin{itemize}
    \item \textbf{Accuracy}: Accuracy measures the proportion of total predictions that the model is correct. It serves as a general measure of the model's overall correctness.
    \item \textbf{Precision}: Precision indicates how many of the instances that the model identified as positive are actually positive. 
    \item \textbf{Recall}: Recall measures the proportion of actual positives that the model correctly identified. 
    \item \textbf{F1-Score}: The F1-score is a weighed average of precision and recall. 
\end{itemize}

\subsection*{Bee Image Object Detection Results}

\begin{figure}[tb]
    \centering
    \includegraphics[width=0.85\textwidth, height=5cm]{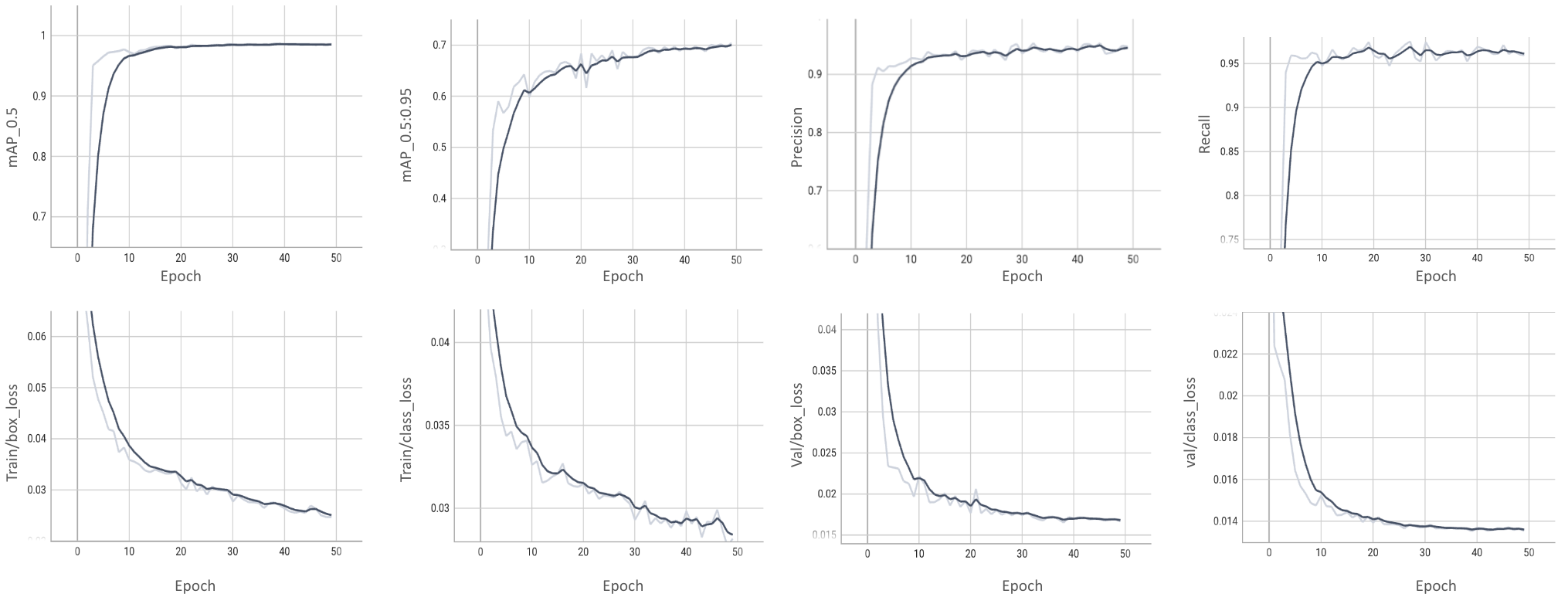} 
    \caption{Summary of YOLO v5 model performance}
    \label{fig:Yolo}
\end{figure}

In the study, a computer vision model, YOLO v5, was used to locate and crop bees within various pictures. As illustrated in Figure \ref{fig:Yolo}, the model demonstrated high accuracy, achieving a 98.6\% mean average precision (mAP@50). The rate was assessed by comparing how closely the locations identified by the model matched the real locations of the bees when at least half of the area overlapped. The accuracy was further tested under varying conditions, where the overlap between the predicted and actual locations of bees ranged from 50\% to 95\%. This thorough assessment demonstrated the model's ability to accurately detect bees in a variety of conditions. The mAP@X is calculated by: \[
\text{mAP@X} = \frac{1}{N} \sum_{i=1}^{N} \text{AP$_i$}
\]

where 
\begin{align*}
& N : \text {the number of classes}\\
& \text{AP$_i$} : \text 
 {the average precision (area under precision vs recall curve) at an IoU threshold of X for each class}
\end{align*}

\noindent During the process, the model's improvement was monitored through loss curves, as shown in Figure \ref{fig:Yolo}. These curves in the training and validation datasets showed similar trends, which indicated that the model could accurately classify bees in new, previously unseen data. The steady progress of these curves in both the training and validation phases suggested that the model was learning effectively. 

\subsection*{Bee Audio Object Detection Results}

\noindent Four one-dimensional CNNs with distinct audio features were applied to identify bee sounds in audio recordings,  with varying accuracies as shown in Table \ref{tab:audio_features}. The model using the Mel Spectrogram, focusing on sound frequency, achieved a 76.41\% accuracy rate, while the model using the Chromagram, analyzing sound pitch, reached 82.39\%. This finding indicated that analyzing pitches is more effective in distinguishing bee sounds from other noises, which is insightful for bee behavior research. \\

\noindent 
Most models performed consistently across different health categories. For example, the model based on Chromagram achieved similar F1-scores of 91.20\% and 96.64\% for both bee and non-bee sounds, as shown in Table \ref{tab:bee_no_bee}. This indicated that the models were equally proficient at recognizing bee and non-bee sounds. Such a balance is crucial when it is equally important to avoid mistakenly classifying non-bee sounds as bee sounds and missing actual bee sounds.

\subsection*{Bee Image Health Assessment Results}

\noindent 
In the study, four deep learning models were developed to classify bee health through image analysis. Despite their different designs, these models showed similar performance in determining the health status of bees. As shown in Table \ref{tab:visual_model}, their accuracies ranged from 65.34\% for Inception v3 to 69.89\% for VGG16.\\

\begin{table}[!h]
    \centering
    \begin{tabular}{lcccc}
        \toprule
        \textbf{Audio Features} & \textbf{Accuracy} & \textbf{Precision} & \textbf{Recall} & \textbf{F1-Score} \\
        \midrule
        Mel Spectrogram & 76.41\% & 83.41\% & 76.41\% & 75.10\% \\
        MFCC & 82.39\% & 82.81\% & 82.39\% & 82.34\% \\
        STFT & 89.44\% & 89.47\% & 89.44\% & 89.43\% \\
        Chromagram & 95.13\% & 95.10\% & 95.13\% & 95.10\% \\
        \bottomrule
    \end{tabular}
    \caption{Summary of four audio deep learning model performances for bee object detection}
    \label{tab:audio_features}
\end{table}

\begin{table}[!h]
    \centering
    \begin{tabular}{lcc}
        \toprule
        \textbf{Audio Features} & \textbf{Bee} & \textbf{No\_Bee} \\
        \midrule
        Mel Spectrogram & 80.80\% & 69.41\% \\
        MFCC & 81.34\% & 83.33\% \\
        STFT & 89.29\% & 89.58\% \\
        Chromagram & 91.20\% & 96.64\% \\
        \bottomrule
    \end{tabular}
    \caption{Summary of F1-scores for bee object detection across four audio deep learning models}
    \label{tab:bee_no_bee}
\end{table}

\begin{table}[bt]
    \centering
    \label{tab:model_performance}
    \begin{tabular}{lcccc}
        \toprule
        \textbf{Models} & \textbf{Accuracy} & \textbf{Precision} & \textbf{Recall} & \textbf{F1-score} \\
        \midrule
        Inception v3 & 65.34\% & 65.43\% & 65.34\% & 64.83\% \\
        MobileNet v2 & 65.91\% & 65.74\% & 65.91\% & 65.72\% \\
        CNN & 68.18\% & 70.30\% & 68.18\% & 67.43\% \\
        VGG16 & 69.89\% & 69.26\% & 69.89\% & 68.89\% \\
        \bottomrule
    \end{tabular}
    \caption{Summary of four visual deep learning model performances for bee health assessment}
    \label{tab:visual_model}
\end{table}

\begin{table}[!hbt]
    \centering    
    \label{tab:health_condition_f1}
    \begin{tabular}{lcccc}
        \toprule
        \textbf{Models} & \textbf{Healthy} & \textbf{Ants Infestation} & \textbf{Missing Queen} & \textbf{Pesticide Exposure} \\
        \midrule
        Inception v3 & 73.87\% & 66.67\% & 69.47\% & 45.07\% \\
        MobileNet v2 & 71.15\% & 65.91\% & 73.68\% & 46.15\% \\
        CNN & 68.24\% & 74.42\% & 74.23\% & 54.76\% \\
        VGG16 & 79.25\% & 74.73\% & 69.66\% & 48.48\% \\
        \bottomrule
    \end{tabular}
    \caption{Summary of F1-scores for each bee health condition across four visual deep learning models}
    \label{tab:visual_class}
\end{table}

\noindent 
One of the main challenges for all the models was identifying bees exposed to pesticides, as this task yielded the lowest F1-scores ranging from 45.07\% to 48.48\%, as illustrated in table \ref{tab:visual_class}. This difficulty may be attributed to the small quantities of pesticides exposed to bees in the data collection process. Another possible reason might be the insensitivity of the images to subtle behavioral changes in bees caused by pesticide exposure.

\subsection*{Bee Audio Health Assessment Results}

\noindent Four advanced deep learning models utilizing STFT audio feature were employed for classifying bee health from sounds. VGG16, known for its ability to detect complex patterns in spectrograms, achieved the highest accuracy at 81.25\%. LSTM, which excels at recognizing patterns over time in audio, followed with an accuracy of 77.84\%. These results are shown in Table \ref{tab:audio_model_performance}.\\

\noindent Both VGG16 and LSTM were more effective than other models in identifying bees affected by pesticides. They increased the detection rates for pesticide-exposed bees by 24.52\% to 59.73\%, as shown by the improved F1-scores in Table \ref{tab:audio_health_condition_f1}. It demonstrated that 
combining these advanced models with sounds could capture subtle cues in bee behavior, particularly in detecting the impacts of environmental stressors like pesticides.

\subsection*{Attention-based Multimodal Neural Network Results}

\noindent An attention-based multimodal neural network (AMNN) was developed to combine both visual and audio information to assess bee health more accurately. The model achieved an accuracy of 92.61\%, significantly outperforming the best visual model at 69.89\% and the best audio model at 81.25\%, as shown in Table \ref{tab:compare}.

\begin{table}[!h]
    \centering

    \begin{tabular}{lcccc}
        \toprule
        \textbf{Models} & \textbf{Accuracy} & \textbf{Precision} & \textbf{Recall} & \textbf{F1-Score} \\
        \midrule
        1D CNN & 68.75\% & 72.74\% & 68.75\% & 68.36\% \\
        2D CNN & 71.59\% & 77.46\% & 71.59\% & 71.22\% \\
        LSTM & 77.84\% & 81.09\% & 77.84\% & 77.71\% \\
        VGG16 & 81.25\% & 81.45\% & 81.25\% & 81.32\% \\
        \bottomrule
    \end{tabular}
    \caption{Summary of four audio deep learning model performances for bee health assessment}
    \label{tab:audio_model_performance}
\end{table}

\begin{table}[!h]
    \centering
    \begin{tabular}{lcccc}
        \toprule
        \textbf{Models} & \textbf{Healthy} & \textbf{Ant Infestation} & \textbf{Missing Queen} & \textbf{Pesticide Exposure} \\
        \midrule
        1D CNN & 62.50\% & 73.58\% & 69.57\% & 66.67\% \\
        2D CNN & 71.26\% & 86.54\% & 66.09\% & 52.17\% \\
        LSTM & 71.03\% & 91.89\% & 62.16\% & 83.33\% \\
        VGG16 & 94.44\% & 76.00\% & 70.33\% & 83.02\% \\
        \bottomrule
    \end{tabular}
    \caption{Summary of F1-scores for each bee health condition across four audio deep learning models}
    \label{tab:audio_health_condition_f1}
\end{table}

\noindent 
An interesting observation was how effectively the visual and audio cues complemented each other. For instance, in cases where visual cues alone misclassified 26.14\% of instances, audio signals correctly identified 86.79\% of them. Conversely, with 14.77\% misclassifications by audio cues, visual information correctly identified 78.79\%. This suggests that integrating these two signal types provided a more complete and accurate picture of bee health.\\

\noindent The AMNN model not only improved accuracy overall but also significantly enhanced the ability to classify specific bee health conditions. For example, when detecting bees affected by pesticides, the model's effectiveness, as measured by the F1-score, increased dramatically from 48.48\% to 90.57\%. Similar improvements were observed in the other three bee health categories, with an increase of around 20.5\% to 30.51\% in F1-scores. As a result, the model proved to be highly reliable, consistently achieving a success rate above 90\% for all types of bee health states. These notable enhancements in model performance were detailed in Table \ref{tab:health_state_f1}.\\

\noindent 
The models using only visual or audio data were quite quick to train, taking approximately 2.5 seconds each. However, the AMNN model, which combined both types of data, took slightly longer, at about 3.68 seconds. Similarly, when making predictions, the visual-only and audio-only models were fast, each taking roughly 0.1 seconds. The AMNN model was slightly slower, taking about 0.16 seconds, as documented in Table \ref{tab:training_inference_times}.\\

\begin{table}[!h]
\centering
\begin{tabular}{lcccccc}
    \toprule
    \textbf{Models} & \textbf{Accuracy} & \textbf{Precision} & \textbf{Recall} & \textbf{F1-score} \\
    \midrule
    Visual-only & 69.89\% & 69.26\% & 69.89\% & 68.89\% \\
    Audio-only & 81.25\% & 81.45\% & 81.25\% & 81.32\% \\
    AMNN       & 92.61\% & 92.82\% & 92.61\% & 92.57\% \\
    \bottomrule
\end{tabular}
\caption{Summary of visual-only, audio-only, and AMNN model performances for bee health assessment}
\label{tab:compare}
\end{table}

\begin{table}[!h]
    \centering
    \begin{tabular}{lcccc}
        \toprule
        \textbf{Models} & \textbf{Healthy} & \textbf{Ant Infestation} & \textbf{Missing Queen} & \textbf{Pesticide Exposure} \\
        \midrule
        Visual-only & 79.25\% & 74.73\% & 69.66\% & 48.48\% \\
        Audio-only & 94.44\% & 76.00\% & 70.33\% & 83.02\% \\
        AMNN & 95.50\% & 92.00\% & 90.91\% & 90.57\% \\
        \bottomrule
    \end{tabular}
    \caption{Summary of F1-scores for each bee health condition across visual-only, audio-only, and AMNN}
    \label{tab:health_state_f1}
\end{table}

\begin{table}[tb]
    \centering
    \begin{tabular}{lcc}
        \toprule
        \textbf{Models} & \textbf{Training Time (s)} & \textbf{Inference Time (s)} \\
        \midrule
        Visual-only & 2.48 & 0.11 \\
        Audio-only & 2.51 & 0.10 \\
        AMNN & 3.68 & 0.16 \\
        \bottomrule
    \end{tabular}
    \caption{Comparison of model processing times across visual-only, audio-only, and AMNN models}  \label{tab:training_inference_times}
\end{table}

\section*{Discussion}
The study presented a holistic and effective beehive monitoring and management system. The system began with bee object detection, in which advanced computer vision and signal processing techniques were utilized to identify bees within images and sound. Subsequently, the AMNN integrated visual and audio features by adaptively assigning different weights based on their importance. This approach outperformed eight existing single-modal ML alternatives without significantly slowing down the model processing time. The AMNN also improved model robustness by consistently achieving high F1-scores for all bee health conditions. \\

\noindent ML models demonstrated their ability to capture distinct bee behaviors through images and sounds in the work. This was evident from the promising results obtained through systematic evaluations of CNNs, Recurrent Neural Networks, and Transfer Learning on bee object detection and health assessment. The model performances were potentially influenced by the training dataset sizes, model structure, and specific audio features. \\

\noindent The integration of visual and audio signals proved to be beneficial, as each signal captured cases that were misclassified by the other. Furthermore, bee sounds were a more reliable behavioral indicator than images, as demonstrated by a 16.25\% superior performance of the audio model when compared to visual models. Several factors could contribute to this observation. Firstly, bees heavily depend on sound for communication, generating vital signals through wing beats, body movements, and vibrations. These sounds convey essential information about food sources, potential threats, and the overall health of the hive. Secondly, environmental conditions, including lighting and obstacles, might obscure visual observations. However, sound can travel through different mediums, offering a more consistent and accurate depiction of bee behavior. This finding highlighted the importance of considering audio data in bee health evaluation.\\

\noindent  \noindent While this study provided valuable insights into the application of ML models for bee monitoring, it was important to acknowledge certain limitations, including limited sample size and restricted data sources. The study primarily focused on Apis mellifera in California, USA, a scope that might not have fully captured the behavioral and environmental variations found in other bee species or locations. Future research could aim for more extensive studies involving a wider variety of bee species, both domesticated and wild, across diverse continents and climates. Collaborating with global researchers, beekeepers, and agricultural organizations would help create more robust datasets. 

\section*{Conclusion}
\noindent To address the critical challenges facing honey bee populations, this study introduces an innovative approach for assessing bee health. The approach seamlessly integrates visual and audio signals. To bridge the gap between beekeeping and advanced technology, a comprehensive system was developed, featuring a user-friendly website for near-real-time monitoring and online evaluation of bee health. Within this system, cameras and microphones were strategically placed around beehives, capturing their activity. The data were then transmitted to an online platform accessible through a Raspberry Pi device. Subsequently, advanced artificial intelligence, as proposed by the research, played a key role in detecting bees and evaluating bee health. The integrated method yielded accuracy improvements of 14\% to 33\% over traditional single signal approaches in bee health assessment. Armed with real-time data and in-depth insights, the solution empowers beekeepers to accurately pinpoint stressors, leading to more targeted beekeeping interventions.

\section*{Data Availability Statement}

There are four datasets used during the study. Two of them have been made publicly accessible on the Kaggle platform. The remaining two datasets are not currently available to the public, due to considerations for intellectual property filings and commercial applications.

\begin{itemize}
    \item \href{https://www.kaggle.com/datasets/andrewlca/bee-image-object-detection}{Bee Image Detection (https://www.kaggle.com/datasets/andrewlca/bee-image-object-detection)}
    \item  \href{https://www.kaggle.com/datasets/andrewlca/bee-audio-object-detection}{Bee Audio Detection (https://www.kaggle.com/datasets/andrewlca/bee-audio-object-detection)}
\end{itemize}

\bibliography{main}

\end{document}